\documentclass[sigconf]{acmart}
\AtBeginDocument{%
  \providecommand\BibTeX{{%
    \normalfont B\kern-0.5em{\scshape i\kern-0.25em b}\kern-0.8em\TeX}}}

\usepackage{hyperref}

\usepackage{hyperxmp}
\usepackage{multirow}
\usepackage{float} 
\usepackage{amsmath}
\usepackage[ruled,linesnumbered]{algorithm2e}
\usepackage{graphicx}
\usepackage{subfigure}
\usepackage{epstopdf}
\usepackage{tablefootnote}
\usepackage[flushleft]{threeparttable}
\usepackage{diagbox}
\usepackage{booktabs}
\usepackage{siunitx}
\usepackage{footmisc}
\usepackage{footnote}
\usepackage{enumitem}
\usepackage[normalem]{ulem}
\useunder{\uline}{\ul}{}

\begin{document}

\title{LexRAG: Benchmarking Retrieval-Augmented Generation in Multi-Turn Legal Consultation Conversation}

\author{Haitao Li}
\authornote{These authors contributed equally to this work.}
\affiliation{%
  \institution{DCST, Tsinghua University\&}
\institution{Quan Cheng Laboratory}
  \city{Beijing}
  \country{China}}
\email{liht22@mails.tsinghua.edu.cn}

\author{Yifan Chen$^*$}
\affiliation{%
  \institution{DCST, Beijing University of Posts and Telecommunications}
  \city{Beijing}
  \country{China}}
\email{chenyifan@bupt.edu.cn}

\author{Yiran Hu$^*$}
\affiliation{%
  \institution{Tsinghua University}
  \city{Beijing}
  \country{China}}
\email{	huyr21@mails.tsinghua.edu.cn}

\author{Qingyao Ai}
\affiliation{%
  \institution{Quan Cheng Laboratory\&}
  \institution{DCST, Tsinghua University}
  \city{Beijing}
  \country{China}}
\email{aiqy@tsinghua.edu.cn}

\author{Junjie Chen}
\affiliation{%
  \institution{Quan Cheng Laboratory\&}
  \institution{DCST, Tsinghua University}
  \city{Beijing}
  \country{China}}
\email{chenjj826@gmail.com}

\author{Xiaoyu Yang}
\affiliation{%
  \institution{Tsinghua University}
  \city{Beijing}
  \country{China}}
\email{y15011462822@163.com}

\author{Jianhui Yang}
\affiliation{%
  \institution{Tsinghua University}
  \city{Beijing}
  \country{China}}
\email{yangjh23@mails.tsinghua.edu.cn}

\author{Yueyue Wu}
\authornote{Corresponding author}
\affiliation{%
  \institution{Quan Cheng Laboratory\&}
  \institution{DCST, Tsinghua University}
  \city{Beijing}
  \country{China}}
\email{wuyueyue@mail.tsinghua.edu.cn}

\author{Zeyang Liu}
\authornote{Corresponding author}
\affiliation{%
\institution{Quan Cheng Laboratory\&}
  \institution{Shandong University}
  \city{Beijing}
  \country{China}}
\email{zeyangliu@sdu.edu.cn}

\author{Yiqun Liu}
\affiliation{%
  \institution{Quan Cheng Laboratory\&}
  \institution{DCST, Tsinghua University}
  \city{Beijing}
  \country{China}}
\email{yiqunliu@tsinghua.edu.cn}
\renewcommand{\shortauthors}{Haitao Li et al.}


\begin{abstract}

Retrieval-augmented generation (RAG) has proven highly effective in improving large language models (LLMs) across various domains. 
However, there is no benchmark specifically designed to assess the effectiveness of RAG in the legal domain, which restricts progress in this area.
To fill this gap, we propose LexRAG, the first benchmark to evaluate RAG systems for multi-turn legal consultations.
LexRAG consists of 1,013 multi-turn dialogue samples and 17,228 candidate legal articles. Each sample is annotated by legal experts and consists of five rounds of progressive questioning.
LexRAG includes two key tasks: (1) Conversational knowledge retrieval, requiring accurate retrieval of relevant legal articles based on multi-turn context. (2) Response generation, focusing on producing legally sound answers. To ensure reliable reproducibility, we develop LexiT, a legal RAG toolkit that provides a comprehensive implementation of RAG system components tailored for the legal domain.
Additionally, we introduce an LLM-as-a-judge evaluation pipeline to enable detailed and effective assessment.
Through experimental analysis of various LLMs and retrieval methods, we reveal the key limitations of existing RAG systems in handling legal consultation conversations.
LexRAG establishes a new benchmark for the practical application of RAG systems in the legal domain, with its code and data available at \url{https://github.com/CSHaitao/LexRAG}.

\end{abstract}
\maketitle
\section{Introduction}

Recently, Retrieval-augmented generation (RAG) has gained significant attention as a powerful approach to improving the performance of large language models (LLMs).
By integrating the strengths of information retrieval with generative models, RAG enables the generation of more accurate, relevant, and contextually appropriate responses based on documents retrieved from up-to-date, reliable sources. While RAG has demonstrated success in various domains, its application in legal domain remains underexplored.

Compared to general domains, RAG faces greater challenges in the legal domain. 
First, legal consultations are more complex, often involving progressively unfolding issues. The users posing questions typically lack sufficient legal knowledge, requiring clarification, confirmation, and correction of details through multiple turns of dialogue. RAG systems must handle irrelevant information from previous interactions and effectively manage abrupt topic shifts. Moreover, in each turn, the relevance of a question to legal knowledge is not simply determined by lexical or semantic similarity~\cite{li2023sailer,li2024delta}. The model needs to consider the context for reasoning, identifying the legal logic and focus of the question to determine the relevant knowledge.

Although some benchmarks have been created to evaluate LLMs in the legal domain, they typically focus on simple tasks, such as legal case retrieval~\cite{10.1145/3626772.3657887,li2024towards} and judgment prediction~\cite{xiao2018cail2018}, failing to capture the complexity that RAG faces in real-world legal scenarios.
To fill this gap, we introduce LexRAG, a benchmark designed for RAG in multi-turn legal consultation conversations.
It consists of 1,013 multi-turn consultation samples and includes 17,228 candidate articles. Each sample comprises five rounds of questions, with responses annotated by legal experts.
In each conversation, the LLM must effectively incorporate previous turns and resolve pronoun references to understand the current query and ensure logical consistency. Additionally, LLMs need to handle abrupt topic shifts, which increase complexity and can degrade retrieval and generation quality as the dialogue history grows.

In LexRAG, we evaluate two key tasks of RAG systems: (1) Conversational Knowledge Retrieval, which assesses the system's ability to retrieve relevant information from a large document corpus based on multi-turn context. (2) Response Generation, which tests its ability to generate accurate, contextually rich answers.
To enable reproducible automated evaluation, we provide an easy-to-use toolkit LexiT, that includes the complete implementation of components for RAG systems in the legal domain. Moreover, we have carefully designed an LLM-as-a-judge evaluation pipeline within the toolkit to enable effective, fine-grained assessment.
We conduct a comprehensive evaluation of various LLMs and retrieval methods, offering an in-depth analysis of the current limitations and shortcomings of RAG systems in the legal domain. Our findings highlight key challenges and suggest future directions for advancing RAG in the legal domain.

In summary, our contributions are three-fold:

\begin{enumerate} 
\item \textbf{First Benchmark for RAG system in Legal Domain.} 
To the best of our knowledge, LexRAG is the first benchmark specifically designed to evaluate RAG in the legal domain. This benchmark provides a standardized platform for evaluating retrieval and generation capabilities in complex legal consultation conversations. It not only advances legal AI technologies but also lays the foundation for the future development of RAG across various domains.

\item \textbf{Open-Source Legal RAG Evaluation Toolkit.} 
In addition to the dataset, we provide LexiT, a dedicated toolkit for RAG in the legal domain.
This toolkit includes various implementations of modules such as processors, retrievers, and generators for RAG systems. Additionally, we have carefully designed the LLM-as-a-judge evaluation framework to enable effective and fine-grained automated assessment.
This contributes to the advancement of research in the legal domain by enabling consistent and comparable evaluations.

\item \textbf{Systematic Evaluation and Analysis.}
Through rigorous evaluation of several LLMs and retrieval methods, we analyze the strengths and limitations of current RAG systems in the legal domain. These observations offer valuable insights and highlight areas for further improvement, providing a roadmap for enhancing RAG-based legal consultation systems in the future.
\end{enumerate}


\section{Related Work}
\label{sec:related}

\subsection{Legal Applications of LLMs}
Large language models (LLMs) have demonstrated strong potential for application in the legal domain~\cite{li2024blade,li2024lexeval}.
Several studies have thoroughly reviewed the current applications of LLMs in the legal domain, highlighting their vast potential in areas like legal consultation and trial assistance~\cite{lai2024large}.
Additionally, researchers have developed LLMs specifically for the legal domain through continued pretraining and fine-tuning~\cite{yue2023disc}.
For example, ChatLaw~\cite{cui2023chatlaw} is built on the Anima-33B model and fine-tuned with a large dataset that includes legal news, statutes, judicial interpretations, legal consultations, exam questions, and court judgments. Meanwhile, LexiLaw~\cite{LexiLaw} is further trained on ChatGLM to offer accurate and reliable legal consultation for legal professionals, students, and the general public. It excels in interpreting legal clauses, analyzing cases, and understanding regulations.

\subsection{Retrieval-Augmented Generation}
Naive LLMs can suffer from hallucinations or provide outdated answers when handling domain-specific tasks or recent information~\cite{tonmoy2024comprehensive,xie2023t2ranking}.
RAG addresses this issue by first retrieving relevant information from external knowledge sources, improving the LLM's accuracy and ensuring timely, up-to-date responses~\cite{fan2024survey,asai2023self}.
The RAG workflow consists of three steps. First, the retriever fetches relevant information from an external knowledge base. Next, the retrieved information is combined with the original query to create an augmented prompt. Then, the generator produces the response based on the augmented prompt.
In recent years, RAG has been widely applied across various fields.  For example, in question-answering systems, LLMs enhance their ability to handle complex queries and generate more accurate responses by integrating a retrieval mechanism.

\subsection{Multi-Turn Conversation}

Dialogue systems are designed to facilitate continuous communication between humans and machines by understanding context and generating coherent responses~\cite{zhou2020kdconv,chen2017survey,ni2023recent}.
These systems are typically classified into task-oriented~\cite{hosseini2020simple} and open-domain systems~\cite{huang2020challenges}. 
Task-oriented systems help users complete specific tasks, such as booking hotels or checking the weather, while open-domain systems engage users in conversations on a wide variety of topics.
The main challenge for these systems is generating coherent and diverse responses to maintain a natural and smooth conversation.
With the development of deep learning and pre-trained models, LLM-based multi-turn dialogue systems have shown excellent performance~\cite{yi2024survey}. These models, pre-trained on large corpora, acquire rich linguistic and world knowledge, enabling them to generate more natural and contextually relevant responses. However, LLMs still face challenges in multi-turn dialogues, such as context understanding, dialogue state tracking, reasoning and planning, and response consistency.

\subsection{Benchmarks in Legal Domian}
In the legal domain, evaluation benchmarks are crucial for the development of LLMs.
Evaluation benchmarks for LLMs are essential for assessing their performance on legal tasks~\cite{li2024lexeval,zhang2024evaluation}.
Researchers have developed benchmarks to evaluate LLMs' capabilities in tasks like legal reasoning, text classification, and question answering.
For example, LexGLUE~\cite{chalkidis2021lexglue} is an English-language benchmark that standardizes the evaluation of models across various legal NLP tasks, such as text classification and case judgment prediction. LexEval introduces a legal cognition taxonomy and organizes 14,150 tasks to systematically evaluate LLMs' abilities in the legal domain.  
Moreover, Li et al.~\cite{li2024legalagentbench} introduced LegalAgentBench, which evaluates LLM agents specifically in the legal domain.

\section{LexRAG}
In this section, we provide a detailed overview of LexRAG, including task definition, characteristics, data construction, and RAG toolkit.

\begin{figure}[t]
\centering
\includegraphics[width=\linewidth]{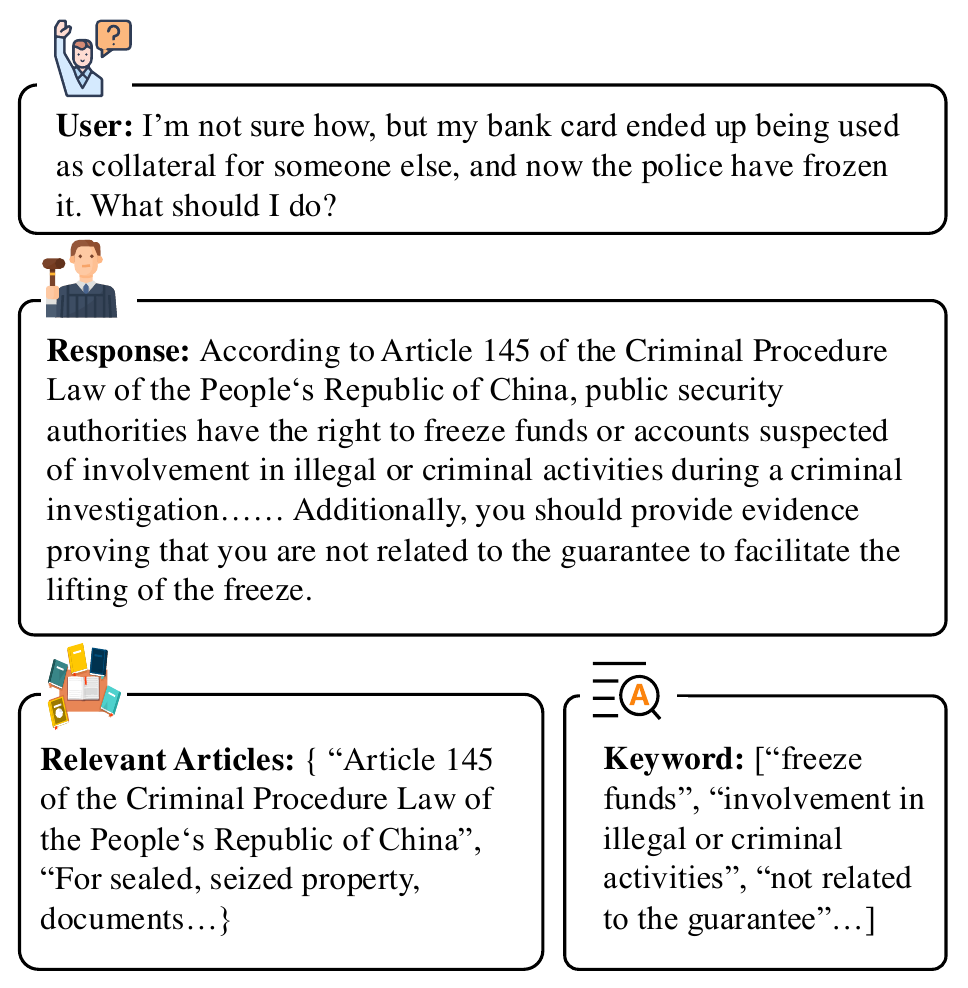}
\caption{An example of a legal consultation in LexRAG.}
\label{figure:example}
\end{figure}

\subsection{Overview}
LexRAG is the first benchmark designed to evaluate the performance of RAG in the legal domain, covering two key tasks: conversational knowledge retrieval and response generation.
The dataset contains 1,013 multi-turn conversations, each with 5 rounds of questions and responses. 
Each conversation is carefully annotated by legal experts to ensure accuracy and professionalism.
Additionally, it includes 17,228 candidate legal articles across various legal domains, such as civil, criminal, contract, and intellectual property law. 
Figure ~\ref{figure:example} illustrates an example of legal consultations in LexRAG.
In addition to the questions and responses, legal experts also identify and annotate relevant legal articles and keywords within the responses. Given that legal terminology has precise meanings, we also use the accuracy of keyword as an evaluation metric.
LexRAG provides a standardized evaluation platform to advance RAG applications in the legal field and support the development of high-quality legal consultation systems.

\subsection{Task Defintion}

LexRAG is designed to evaluate two fundamental tasks: (1) \textbf{Conversational Knowledge Retrieval} and (2) \textbf{Response Generation}.
Compared to general domains, both tasks present unique challenges inherent to the legal domain. 

For conversational knowledge retrieval task, the RAG system must identify legal articles relevant to the current query while considering the context. Formally, given a multi-turn legal dialogue history $H=\{q_1, r_1, ..., q_t\}$,
where $q_t$ represents the user’s question and $r_t$ is the response at turn $t$.
The objective is to retrieve a set of relevant legal articles $A_t = \{a_1,...,a_n\}$ from a predefined legal corpus $\mathcal{D}$.
The retrieved articles should provide authoritative references for answering $q_t$.
Unlike web search tasks that primarily rely on keyword matching or semantic similarity, this task in the legal domain introduces additional complexities. The retrieval models must not only understand the explicit query but also deduce the implicit legal intent behind it. For example, a user might ask a seemingly simple everyday question, but the final answer may involve referencing a complex series of interrelated statutes.
Therefore, the retrieval system must go beyond simple keyword matching and instead focus on a nuanced understanding of legal concepts and relationships.

For Response Generation task, the LLM needs to generate contextually coherent and legally accurate response $a_t$ based on the dialogue history $H$ and retrieved legal articles $A_t$.
In addition to the inherent challenges of multi-turn dialogues, such as anaphora resolution, context dependency, and topic shifts, this task requires LLMs to accurately interpret the legal requirements embedded in the query and apply the retrieved legal information precisely.
In summary, LexRAG requires a deep integration of legal knowledge, multi-turn dialogue management, and advanced retrieval mechanisms.

\subsection{Characteristics}
LexRAG is designed as a comprehensive and reliable benchmark with the following key characteristics:

\textbf{Legal Expertise.}
All responses in LexRAG are carefully annotated and reviewed by experienced legal experts to ensure accuracy and reliability. Additionally, the seed questions are sourced from legal consultation platforms, reflecting real-world legal practices.

\textbf{Multi-Turn.}
In LexRAG, each conversation consists of five interactive turns. User queries often involve anaphora resolution, clarification, and topic shifts. This requires the system to effectively track conversation history and adapt to the evolving legal context.

\textbf{Diversity.} 
LexRAG covers a broad range of real-world legal issues, including 27 query types such as traffic accidents, personal injury, and debt disputes. The retrieval corpus includes 17,728 legal provisions from 222 statutes and regulations, ensuring comprehensive legal coverage for thorough evaluation.

\textbf{Citation-Based Grounding}
A key feature of LexRAG is its focus on legal citation. Most responses explicitly reference legal articles, ensuring alignment with authoritative sources. This approach enhances transparency, verifiability, and highlights the importance of accurate knowledge retrieval in legal consultation.

\subsection{Data Construction}
In this section, we introduce the construction process of LexRAG, including data sources, preprocessing, human annotation, and data analysis.

\subsubsection{Data Source and Preprocessing}
To construct LexRAG, we collected 222 commonly used legal statutes in China, ensuring each was from its latest version. We standardized the formatting of legal provisions and created a structured retrieval corpus with 17,228 legal articles.
Then, we collect seed questions to guide human annotators in structuring and annotating the conversation. These questions are sourced from real-world legal consultation platforms~\footnote{https://www.12348.gov.cn/} to ensure relevance and authenticity. We thoroughly review and exclude queries containing personal information, sensitive content, or legally irrelevant inquiries.

\subsubsection{Human Annotation}
Our annotation team consists of 11 legal experts from China, all of whom have passed the Chinese Judicial Examination and possess extensive legal experience. The team includes six males and five females. Before starting the annotation process, we signed legally binding agreements with all members to ensure compliance with legal standards and protect their rights.

\textbf{Training.}
To ensure dataset quality, we provided systematic training for all legal experts before annotation. We developed a comprehensive annotation guideline, clearly defining the annotation standards and procedures. Additionally, we provided 10 examples to facilitate a better understanding of the annotation requirements.
Each annotator was required to complete 10 pilot tasks and receive feedback and guidance from senior legal experts, who are the creators of the annotation guidelines.
Only those who achieved a pass rate above 90\% were permitted to proceed to the formal annotation phase.

\textbf{Annotation.}
The annotation process begins with an initial seed question. In the subsequent turns, annotators are encouraged to naturally expand the conversation, ensuring that new questions logically follow the existing conversation threads. 

To support the annotation process, we provide a convenient annotation toolkit to annotators. This toolkit uses the BM25~\cite{robertson2009probabilistic} algorithm to retrieve 30 legal articles relevant to the current question from the legal corpus, providing annotators with valuable references. Additionally, annotators have direct access to the full legal corpus, allowing them to manually select the most relevant legal articles for each question.
Then, annotators must provide detailed responses based on their legal expertise. They are also required to highlight keywords in their responses and annotate them with the corresponding legal articles for review and analysis.
To reduce the annotation workload, we use GPT-4o-mini to pre-generate 10 rounds of derivative questions from the initial seed question, covering different perspectives. These generated questions serve as examples, providing inspiration for annotators. To ensure the diversity and originality of the dataset, direct copying is strictly prohibited.

\textbf{Review.}
We implemented a thorough review process to ensure the quality and reliability of the annotated data.
Our gold annotators, who created the annotation guidelines, performed cross-validation of each annotation from multiple perspectives.
Specifically, they evaluated whether the questions were logically coherent and legally valid, whether the responses were accurate and aligned with legal principles, whether the cited legal articles were relevant and correctly referenced, and whether key terms were appropriately annotated.
Any annotations that did not meet the required standards were reviewed by a senior legal expert to ensure they followed legal standards and best practices.
If any issues were found, the data point was sent back for revision and clarification. This process continued until both annotators agreed. Only high-quality annotations were included in the final dataset.
To fairly reward annotators for their expertise, we paid \$0.42 per validated question-response pair. With 5,065 dialogues created, the total payment amounted to \$2,110.

\subsubsection{Annotation Guideline}
To ensure the quality, consistency, and reliability of LexRAG, we have implemented a rigorous validation and annotation process based on the following principles and standards.
Specifically, the annotators follow the annotation pattern of ``parsing the question--identifying relevant legal articles--generating answers--formulate new questions--simulate real-life scenarios''.

\begin{itemize}
\item \textbf{Parsing the Question.}
The seed questions in LexRAG are sourced from real-world legal consultation platforms, meaning they often focus on real-life issues rather than legal facts.
As a result, directly answering the questions may lead to inaccurate responses or a failure to capture the true intent behind the query. To address this, annotators first parse the real-life issues into key legal terms. For example, if a user asks, ``My girlfriend was already pregnant when we were together! Can the child be registered in the household?'' 
The annotator can derive legal terms such as ``household registration'', ``birth certificate'' and ``child out of wedlock'' based on their legal knowledge. These terms are then used to guide the retrieval of relevant legal articles.

\item \textbf{Identifying Relevant Legal Articles.}
Based on the legal terms derived in the previous step, annotators can use our provided retrieval toolkit or keyword matching to identify relevant legal articles from the candidate database. For example, for the question above, the most relevant provision is Article 7 of the ``Household Registration Regulations of the People's Republic of China''.

\item \textbf{Generating Answers.}
Based on the legal logic of syllogism, 
annotators are encouraged to respond by referencing relevant legal articles.
For example, for the question above, the generated response is:  ``According to Article 7 of the `Household Registration Regulations of the People's Republic of China,' children have the right to register for household registration regardless of whether they were born within or outside of marriage. As long as the child's birth complies with national birth policies and relevant supporting documents (such as the birth certificate, parents' ID cards, and household registration book) are provided, the child can legally be registered. The legitimacy of household registration is not affected, even for children born out of wedlock.''

\item \textbf{Formulate New Questions.}
Due to the difficulty of obtaining real-world multi-turn consultation dialogues, especially those involving personal privacy, annotators are encouraged to expand the questions as much as possible. 
We also use GPT-4o-mini to generate questions from different perspectives, serving as a reference.
These questions are intended to inspire annotators and must not be used directly.
For example, based on the previous question, the next query could be: ``What documents are needed for household registration?''

\item \textbf{Simulate Real-life Scenarios.}
Finally, annotators need to modify the questions to better align with real-life scenarios, including replacing nouns with pronouns and making the language more conversational. For example, the question ``What documents are needed for household registration?'' can be rephrased as ``What documents are needed to register the his household?''

\end{itemize}

\begin{table}[t]
\caption{Basic statistic of LexRAG.}
\begin{tabular}{cc}
\hline
\textbf{Statistic}               & \multicolumn{1}{l}{\textbf{\#Number}} \\ \hline
Total Conversations              & 1,013                                 \\
Total Queries                    & 5,065                                 \\
Total Legal Articles             & 17,728                                \\
Avg. Query Length                & 19.43                                 \\
Avg. Response Length             & 165.92                                \\
Avg. Relevant Articles per Query & 1.09                                  \\
Avg. Keywords per Query          & 3.57                                  \\ \hline
\end{tabular}
\label{table: statis}
\end{table}

\begin{figure}[t]
\centering
\includegraphics[width=0.8\linewidth]{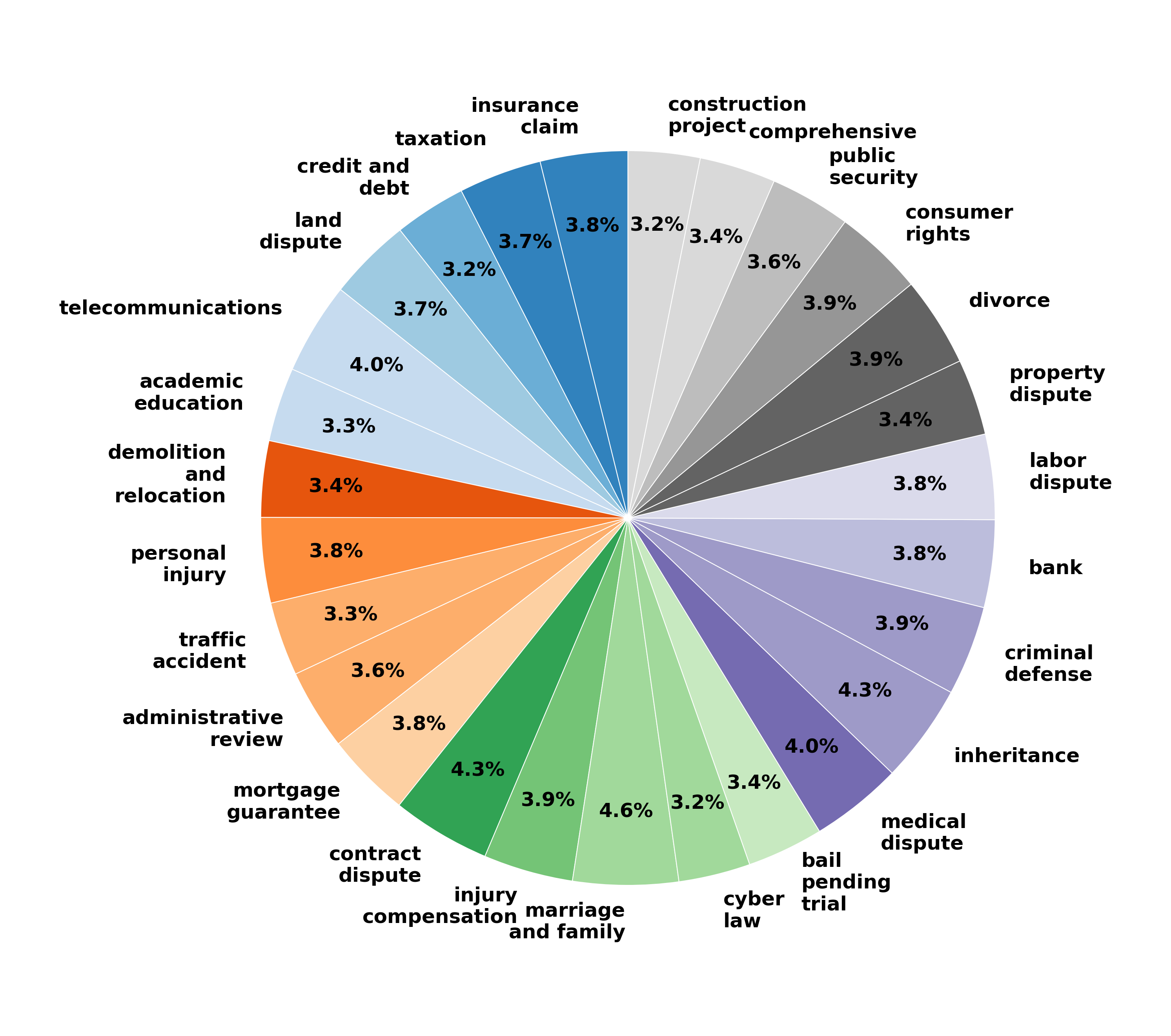}
\vspace{-5mm}
\caption{The distribution of query types.}
\label{figure:task}
\end{figure}

When annotators encounter uncertainties during the annotation process, they should refer to relevant authoritative legal documents, terminology glossaries, or consult legal experts directly to clarify any ambiguities. All decisions made during the annotation process, particularly those following expert consultation, must be clearly documented. This ensures transparency and traceability of decisions, providing a basis for future reviews or revisions and maintaining consistency and standardization.
We encourage annotators to actively provide feedback, propose suggestions for improving the annotation process, or elaborate on any challenges encountered during annotation. The annotation guidelines will be regularly reviewed and updated based on this feedback to meet the evolving needs of LexRAG and enhance annotation quality.

\subsection{Data Analysis}
After careful manual review, LexRAG ultimately contains 1,013 multi-turn conversations, each with 5 interaction rounds. Table ~\ref{table: statis} presents the basic statistics of LexRAG. 
The average response length is notably longer than the query length, suggesting that user queries are typically brief, while responses are designed to offer more comprehensive and detailed information.

As shown in Figure~\ref{figure:task}, LexRAG contains 27 distinct conversation types, which are determined by the seed questions.
We observe that the questions are fairly evenly distributed across these types, indicating that LexRAG covers a wide range of legal domains and exhibits diversity. Overall, LexRAG provides a rich and representative sample for evaluating retrieval and generation capabilities in the legal domain.

\begin{figure}[t]
\centering
\includegraphics[width=\linewidth]{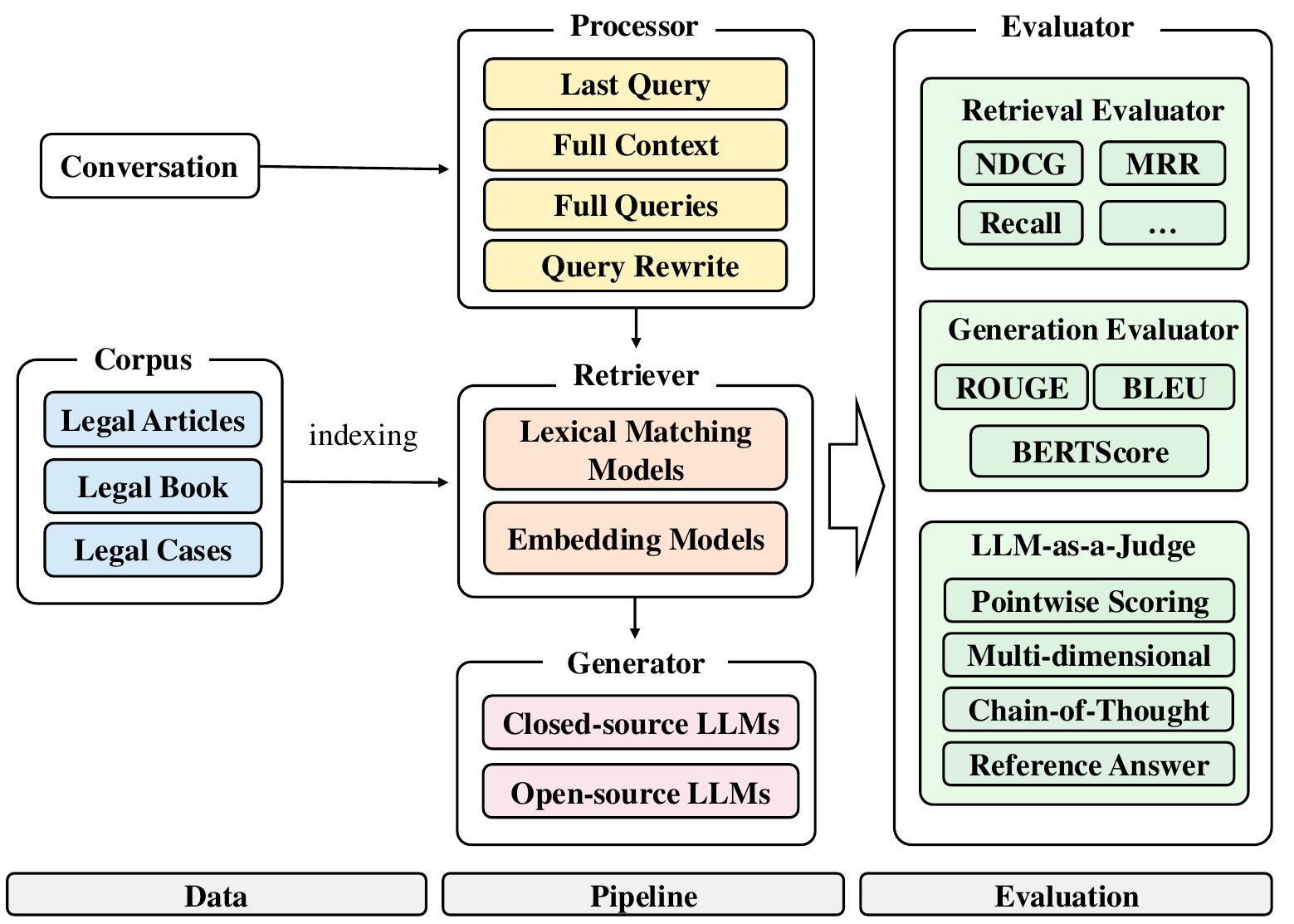}
\caption{Overview of LexiT Components.}
\label{figure:tool}
\end{figure}

\section{LexiT}
To advance RAG system research in the legal domain, we’ve proposed LexiT, a modular and scalable RAG toolkit for legal researchers.
Although there are some general-domain RAG toolkits available, they do not support multi-turn conversations and evaluations tailored to the legal domain~\cite{jin2024flashragmodulartoolkitefficient}.
As shown in Figure \ref{figure:tool}, LexiT consists of three components: Data, Pipeline, and Evaluation.
It integrates all elements of the RAG process into a unified framework and supports standalone applications. This modular design enhances flexibility and allows for high customizability in evaluating different legal scenarios.

\begin{table*}[h]
\caption{The Prompt Template used in LLM-as-a-judge. }
\vspace{-3mm}
\label{table:prompt}
\begin{tabular}{ll}
\hline
\multicolumn{2}{l}{\begin{tabular}[c]{@{\ }p{0.9\textwidth}@{\ }}You are an experienced legal expert responsible for evaluating the quality of legal consultation responses. As an impartial and rigorous evaluator, please assess the AI assistant's response objectively. You will evaluate based on the following five key dimensions:\\ \textbf{Factuality:} Whether the information provided in the response is accurate, based on reliable facts and legal texts.\\ \textbf{User Satisfaction:} Whether the response meets the user’s question and needs, and provides a comprehensive and appropriate answer to the question.\\ \textbf{Clarity:} Whether the response is clear and understandable, and whether it uses concise language and structure so that the user can easily understand it.\\ \textbf{Logical Coherence:} Whether the response maintains overall consistency and logical coherence between different sections, avoiding self-contradiction.\\ \textbf{Completeness:} Whether the response provides sufficient information and details to meet the user’s needs, and whether it avoids omitting important aspects.\\ Longer responses are not necessarily better. The ideal response is short while still meeting the above requirements.\\ \\ You will be provided with the user's multi-turn conversation, a reference answer, and the AI assistant's response to the final question in the conversation. When starting your evaluation, please follow these steps:\\ 1.  Compare the AI assistant's response with the reference answer, highlighting shortcomings and providing further explanations.\\ 2.  Evaluate each dimension strictly according to the scoring criteria outlined above. All dimensions must adhere to the high standard of the reference answer, avoiding inflated scores.\\ 3. Combine the evaluations from all dimensions to assign an overall score between 1 and 10. The final score should reflect the overall performance across all dimensions and not be unduly influenced by a single strength.\\ 4. Provide strict and consistent scoring, following the rules below. In general, the higher the quality of the model’s response, the higher the score.\\ \\ Scoring Stardards:\\ 1-2 points: The model provides severe factual errors, incorrect or irrelevant legal texts and interpretations, or completely unrelated responses. The language is confusing, overly long, or incomprehensible, and the structure is extremely complex, causing user confusion. The answer lacks logical coherence, with incoherent reasoning and contradictions, and fails to provide any valid information. Key details are missing.\\ ......\\ As an example, the reference answer can score 8 points.\\ \\ Please provide a detailed evaluation for each dimension, followed by the corresponding score. All scores should be integers. The final evaluation should be returned in the following format.\\
......

\end{tabular}} \\ \hline
\end{tabular}
\vspace{-3mm}
\end{table*}

\textbf{Data.} 
The data component consists of two key elements: input conversations and corpora.
The conversation format can be either single-turn or multi-turn.
Single-turn conversations are simple QA dialog, while multi-turn conversations provide previous dialogue history as context.
For the corpora, we collect raw data from three different sources. 
In addition to Legal Articles, which serve as the candidate corpus in this paper, Legal Books and Legal Cases are also included in the toolkit for researchers' convenience. Specifically, Legal Articles contains 17,228 provisions from various Chinese statutory laws. Legal Book refers to the National Unified Legal Professional Qualification Examination Counseling Book, which consists of 15 topics and 215 chapters, totaling 26,951 provisions. Legal Cases includes 2,370 officially published guiding cases in China. In the future, we plan to expand the corpus with more legal data.

\textbf{Pipeline.} 
The pipeline component consists of processor, retriever, and generator.
The processor is responsible for converting the conversation into queries used by the retriever.
There are several strategies for constructing the query, including using the last question, the entire conversation context, or the entire query history. 
Moreover, we also predefined a query rewrite strategy, which employs an LLM to integrate all necessary context into a clear, standalone question. Users can easily customize the preprocessing strategy by inheriting and modifying the relevant classes.
For the retriever, we integrate various popular retrieval methods. For lexical matching, we use the Pyserini~\cite{lin2021pyserinieasytousepythontoolkit} library to implement BM25~\cite{robertson2009probabilistic} and QLD~\cite{zhai2008statistical}. For dense retrieval, we support advanced models such as BGE~\cite{chen2024bge} and GTE. Users can encode vectors using locally loaded models or API calls. We employ the Faiss~\cite{johnson2019billion} for index construction, ensuring compatibility with mainstream indexing formats.
In the generator module, we leverage vLLM~\cite{kwon2023efficient} and Huggingface~\footnote{\url{https://huggingface.co}} to support mainstream LLMs. LexiT also supports flexible prompt customization by combining queries with retrieved content, enabling users to easily adjust generation strategies.

\begin{table*}[t]
\centering
\vspace{-3mm}
\caption{Retrieval Performance of different methods on LexRAG using Recall(\%) and nDCG(\%) metrics. The best results are highlighted in bold.}
\vspace{-3mm}
\begin{tabular}{llcccccccc}
\hline
Retriever        & Processor     & Recall@1       & Recall@3       & Recall@5       & Recall@10      & NDCG@1         & NDCG@3         & NDCG@5         & NDCG@10        \\ \hline
BM25             & Last Query    & 5.64           & 10.60          & 13.80          & 18.75          & 6.13           & 9.11           & 10.52          & 12.21          \\
BM25             & Full Context  & 4.89           & \textbf{11.20} & \textbf{15.02} & \textbf{21.28} & 5.31           & 8.92           & 10.58          & \textbf{12.70} \\
BM25             & Full Queries  & 3.82           & 7.89           & 11.86          & 17.86          & 4.15           & 6.58           & 8.30           & 10.36          \\
BM25             & Query Rewrite & \textbf{5.73}  & 10.95          & 14.13          & 18.84          & \textbf{6.21}  & \textbf{9.35}  & \textbf{10.74} & 12.36          \\ \hline
BGE-base         & Last Query    & 9.86           & \textbf{19.26} & 24.40          & 31.41          & 10.70          & \textbf{16.22} & 18.46          & 20.84          \\
BGE-base         & Full Context  & 6.04           & 13.09          & 17.48          & 25.26          & 6.55           & 10.61          & 12.50          & 15.17          \\
BGE-base         & Full Queries  & 5.40           & 12.15          & 16.64          & 24.22          & 5.86           & 9.75           & 11.72          & 14.32          \\
BGE-base         & Query Rewrite & \textbf{9.89}  & 19.19          & \textbf{24.46} & \textbf{31.66} & \textbf{10.74} & 16.17          & \textbf{18.48} & \textbf{20.92} \\ \hline
GTE-Qwen2-1.5B              & Last Query    & 11.37          & 21.35          & 26.55          & 33.13          & 12.34          & 18.23          & 20.46          & 22.68          \\
GTE-Qwen2-1.5B              & Full Context  & 7.98           & 16.42          & 21.77          & 29.93          & 8.67           & 13.45          & 15.72          & 18.45          \\
GTE-Qwen2-1.5B              & Full Queries  & 7.11           & 15.26          & 20.19          & 27.71          & 7.72           & 12.47          & 14.58          & 17.10          \\
GTE-Qwen2-1.5B              & Query Rewrite & \textbf{11.46} & \textbf{21.37} & \textbf{26.60} & \textbf{33.33} & \textbf{12.44} & \textbf{18.29} & \textbf{20.53} & \textbf{22.81} \\ \hline
text-embedding-3 & Last Query    & 10.07          & \textbf{18.02} & 21.91          & 27.84          & 10.94          & 15.56          & 17.25          & 19.26          \\
text-embedding-3 & Full Context  & 8.80           & 17.46          & \textbf{22.80} & 30.71          & 9.56           & 14.53          & 16.81          & \textbf{19.49} \\
text-embedding-3 & Full Queries  & 6.89           & 13.86          & 18.29          & 24.75          & 7.48           & 11.46          & 13.38          & 15.55          \\
text-embedding-3 & Query Rewrite & \textbf{10.20} & 17.97          & 21.97          & 28.08          & \textbf{11.08} & \textbf{15.58} & \textbf{17.30} & 19.39          \\ \hline
\end{tabular}
\label{retrieval}
\end{table*}

\textbf{Evaluation.}
The evaluation module consists of three key components: the retrieval evaluator, the generation evaluator, and the LLM-as-a-judge.
The retrieval evaluator assesses the relevance and accuracy of retrieved documents, supporting the calculation of mainstream automated metrics such as NDCG~\cite{wang2013theoretical}, Recall, MRR~\cite{worster2004advanced}, Precision, and F1. The generation evaluator measures the consistency between generated responses and reference answers, supporting automated metrics like ROUGE~\cite{lin2004rouge}, BLEU~\cite{papineni2002bleu}, METEOR~\cite{banerjee2005meteor}, and BERTScore~\cite{zhang2019bertscore}.

While current automated metrics are useful, they often fail to capture key aspects such as fluency, logical coherence, and factuality, making it difficult to meet the demands of multi-dimensional evaluation criteria.
 Human evaluation, often considered the gold standard, is time-consuming and labor-intensive, making large-scale assessments difficult.
Therefore, we introduce LLM-as-a-judge to enable efficient multi-dimensional automated evaluation.
As LLM capabilities continue to advance, they have been widely adopted as evaluators, demonstrating high consistency with human assessments~\cite{li2024llms,chu2024prepeerreviewbased,li2024calibraeval}. However, evaluating legal texts remains particularly challenging due to the need for a deep understanding of legal nuances and complex reasoning.
To overcome this, we carefully designed the LLM judge evaluation framework within our toolkit to ensure the professionalism and reliability of legal text assessments.

As shown in Figure ~\ref{figure:tool}, the LLM-as-a-judge has four key features:

\begin{itemize}[leftmargin=*]
\item \textbf{Pointwise Scoring.} 
We use a pointwise scoring method due to its enhanced flexibility and scalability.
Specifically, the LLM judge assigns a score from 1 to 10 to each response, considering the dialogue context, the current question, and the reference answer. This method enables a more detailed evaluation of each response while ensuring consistency across the same criteria.

\item \textbf{Multi-dimensional Evaluation.}
Inspired by Wang et al.~\cite{wang2024user},  we develop five evaluation dimensions: Factuality, User Satisfaction, Clarity, Logical Coherence, and Completeness, each with detailed explanations and scoring standards.
We also remind the LLM judges that longer responses are not always better, to mitigate potential biases.

\item \textbf{Chain-of-Thought Reasoning.}
To obtain more reliable evaluation results, the LLM-as-a-judge evaluation framework incorporates chain-of-thought reasoning~\cite{wei2022chain}.
Specifically, LLM judges first compare the generated response with the reference answer, identify shortcomings, and provide further explanations. Then, they evaluate each dimension based on the established scoring criteria. Finally, the LLM judges combine the evaluations from all dimensions to generate an overall score.

\item \textbf{Reference-based Evaluation.}
Due to the specialized knowledge required for legal evaluations, we provide the LLM judges with human expert-annotated responses as references. These reference answers serve as a baseline, with a score of 8 representing the standard for a well-constructed answer.

\end{itemize}

In Table \ref{table:prompt}, we provide the prompt template used in LLM-as-a-judge, which includes the evaluation criteria, chain-of-thought process, scoring standards, and output format requirements.

\section{Conversational Knowledge Retrieval}
In this section, we evaluate the performance of different processing strategies and retrieval models in LexRAG.

\subsection{Experimental Setting}
We evaluate several popular retrieval models, including BM25~\cite{robertson2009probabilistic}, BGE-base-zh~\cite{chen2024bge}, GTE-Qwen2-1.5B-instruct~\footnote{\url{https://huggingface.co/Alibaba-NLP/gte-Qwen2-1.5B-instruct}}, and text-embedding-3-small~\footnote{\url{https://platform.openai.com/docs/guides/embeddings}}. 
These models cover lexical matching and dense retrieval techniques, making them representative.
We report commonly used evaluation metrics including Recall and nDCG, evaluated at positions @1, @3, @5 and @10.

For the processor, we test four different strategies.

\begin{itemize}[leftmargin=*]
\item \textbf{Last Query.}  Using the last query in the conversation as input to the retriever.
\item \textbf{Full Context.}  Using the entire conversation as input to the retriever.
\item \textbf{Full Queries.}  Using all queries in the conversation as input to the retriever.
\item \textbf{Query Rewrite.} Using GPT-4o-mini to turn the relevant context into a clear, standalone question. Specific prompts and examples can be found on our GitHub.
\end{itemize}

\subsection{Retrieval Result}

\begin{table*}[t]
\caption{The Accuracy and LLM judge score of different baselines on LexRAG. The best results are highlighted in bold.}
\small
\begin{tabular}{ll|cccccccccccc}
\hline
\multirow{2}{*}{Model} & \multirow{2}{*}{type} & \multicolumn{2}{c}{1-turn}      & \multicolumn{2}{c}{2-turn}      & \multicolumn{2}{c}{3-turn}      & \multicolumn{2}{c}{4-turn}      & \multicolumn{2}{c}{5-turn}      & \multicolumn{2}{c}{ALL}         \\
                       &                       & Accuracy        & LLM           & Accuracy        & LLM           & Accuracy        & LLM           & Accuracy        & LLM           & Accuracy        & LLM           & Accuracy        & LLM           \\ \hline
GLM-4-Flash            & Zero                  & 0.3431          & 6.11          & 0.3534          & 6.86          & 0.3738          & 6.87          & 0.3737          & 6.88          & 0.3726          & 6.82          & 0.3633          & 6.71          \\
GLM-4-Flash            & Retriever             & 0.3403          & 5.92          & 0.3670          & 6.75          & 0.3783          & 6.78          & 0.3794          & 6.83          & 0.3820          & 6.77          & 0.3694          & 6.61          \\
GLM-4-Flash            & Reference             & \textbf{0.5843} & \textbf{6.52} & \textbf{0.4776} & \textbf{7.06} & \textbf{0.4610} & \textbf{6.99} & \textbf{0.4451} & \textbf{6.93} & \textbf{0.4382} & \textbf{6.89} & \textbf{0.4812} & \textbf{6.88} \\ \hline
GLM-4                  & Zero                  & 0.3468          & 6.40          & 0.3462          & 7.08          & 0.3782          & 7.13          & 0.3809          & 7.15          & 0.3836          & 7.16          & 0.3671          & 6.98          \\
GLM-4                  & Retriever             & 0.3713          & 6.24          & 0.3726          & 6.87          & 0.3981          & 6.90          & 0.3934          & 6.92          & 0.3905          & 6.88          & 0.3851          & 6.76          \\
GLM-4                  & Reference             & \textbf{0.6151} & \textbf{6.76} & \textbf{0.5423} & \textbf{7.27} & \textbf{0.5208} & \textbf{7.30} & \textbf{0.4906} & \textbf{7.27} & \textbf{0.4862} & \textbf{7.25} & \textbf{0.5310} & \textbf{7.17} \\ \hline
GPT-3.5-turbo          & Zero                  & 0.3016          & 6.10          & 0.3032          & 6.63          & 0.3173          & 6.54          & 0.3218          & 6.47          & 0.3335          & 6.49          & 0.3154          & 6.45          \\
GPT-3.5-turbo          & Retriever             & 0.3217          & 5.88          & 0.3057          & 6.41          & 0.3220          & 6.38          & 0.3278          & 6.31          & 0.3231          & 6.30          & 0.3200          & 6.26          \\
GPT-3.5-turbo          & Reference             & \textbf{0.5063} & \textbf{6.53} & \textbf{0.4055} & \textbf{6.90} & \textbf{0.3970} & \textbf{6.74} & \textbf{0.3862} & \textbf{6.63} & \textbf{0.3946} & \textbf{6.65} & \textbf{0.4179} & \textbf{6.69} \\ \hline
GPT-4o-mini            & Zero                  & 0.2982          & 5.95          & 0.2962          & 6.48          & 0.3195          & 6.39          & 0.3075          & 6.28          & 0.3219          & 6.27          & 0.3086          & 6.28          \\
GPT-4o-mini            & Retriever             & 0.3308          & 5.92          & 0.3395          & 6.51          & 0.3411          & 6.38          & 0.3445          & 6.32          & 0.3468          & 6.33          & 0.3405          & 6.29          \\
GPT-4o-mini            & Reference             & \textbf{0.5249} & \textbf{6.39} & \textbf{0.4265} & \textbf{6.83} & \textbf{0.4063} & \textbf{6.62} & \textbf{0.3948} & \textbf{6.47} & \textbf{0.3953} & \textbf{6.48} & \textbf{0.4295} & \textbf{6.56} \\ \hline
Qwen-2.5-72B           & Zero                  & 0.3583          & 6.83          & 0.4037          & 7.37          & 0.4260          & 7.32          & 0.4271          & 7.33          & 0.4266          & 7.33          & 0.4083          & 7.24          \\
Qwen-2.5-72B           & Retriever             & 0.3723          & 6.46          & 0.4097          & 7.24          & 0.4296          & 7.23          & 0.4249          & 7.27          & 0.4359          & 7.28          & 0.4144          & 7.09          \\
Qwen-2.5-72B           & Reference             & \textbf{0.6045} & \textbf{7.14} & \textbf{0.5260} & \textbf{7.45} & \textbf{0.5186} & \textbf{7.49} & \textbf{0.5117} & \textbf{7.41} & \textbf{0.5015} & \textbf{7.37} & \textbf{0.5324} & \textbf{7.37} \\ \hline
Llama-3.3-70B          & Zero                  & 0.2556          & 4.98          & 0.2695          & 5.63          & 0.2846          & 5.46          & 0.2800          & 5.21          & 0.2894          & 5.22          & 0.2758          & 5.30          \\
Llama-3.3-70B          & Retriever             & 0.2735          & 5.26          & 0.2755          & 5.69          & 0.2861          & 5.47          & 0.2850          & 5.32          & 0.2884          & 5.18          & 0.2817          & 5.38          \\
Llama-3.3-70B          & Reference             & \textbf{0.5468} & \textbf{5.83} & \textbf{0.4583} & \textbf{6.30} & \textbf{0.4459} & \textbf{6.02} & \textbf{0.4423} & \textbf{5.85} & \textbf{0.4454} & \textbf{5.83} & \textbf{0.4677} & \textbf{5.97} \\ \hline
Claude-3.5-sonnet      & Zero                  & 0.2464          & 5.60          & 0.2856          & 6.03          & 0.2989          & 5.95          & 0.305           & 5.86          & 0.3064          & 5.91          & 0.2884          & 5.87          \\
Claude-3.5-sonnet      & Retriever             & 0.3667          & \textbf{6.42} & 0.3436          & \textbf{6.90} & 0.3554          & \textbf{6.88} & 0.3604          & \textbf{6.79} & 0.3597          & \textbf{6.77} & 0.3571          & \textbf{6.75} \\
Claude-3.5-sonnet      & Reference             & \textbf{0.5030} & 6.26          & \textbf{0.4304} & 6.60          & \textbf{0.4039} & 6.32          & \textbf{0.3786} & 6.12          & \textbf{0.3840} & 6.19          & \textbf{0.4199} & 6.30          \\ \hline
\end{tabular}
\label{genera}
\end{table*}

Table \ref{retrieval} presents the performance of different retrieval models and processing strategies on LexRAG. Based on the experimental results, we draw the following conclusions:

\begin{itemize}[leftmargin=*]
\item \textbf{Comparing Different Retrieval Models.}  
Dense retrieval methods outperform traditional lexical matching methods like BM25.
Overall, GTE-Qwen2-1.5B-instruct achieved the best results. 
This can be attributed to the challenge that queries in multi-turn consultations often involve pronouns, making basic lexical matching insufficient for identifying relevant legal articles.

\item \textbf{Comparing Different Process Strategies.}  
For dense retrieval methods, the query rewrite strategy typically produces the best results.  This is likely because it integrates relevant information while minimizing the influence of irrelevant data. Moreover, the last query strategy performs better than using all queries or all contexts. 
We speculate that this is due to the inclusion of previous conversation content without filtering, which may introduce noise and distort the query's semantics, ultimately reducing performance.
For lexical matching models, such as BM25, the full context strategy generally achieves the best recall results.
This is likely because providing more context helps reduce the ambiguity caused by pronouns and other context-dependent terms, improving the retrieval of relevant legal articles. Given these findings, we recommend adjusting processing strategies to align with the strengths of each retrieval method, ensuring optimal performance in different scenarios.

\item  \textbf{Existing LLMs Still Struggle with Conversation Knowledge Retrieval in the Legal Domain.}
Overall, current methods perform suboptimally in conversational knowledge retrieval task. Even with the best combination of model and processing strategy, the highest achieved Recall@10 is only 33.33\%.
This result highlights the challenging of LexRAG, demonstrating that existing retrieval models struggle to effectively handle the nuances of legal consultations.
This gap presents an opportunity for the community to create more specialized models that can better address the unique challenges posed by legal contexts.
\end{itemize}

\section{Response Generation}
In this section, we report the performance of different LLMs in response generation task.

\subsection{Experimental Setting}

We evaluated several popular models: GLM-4-flash~\cite{glm2024chatglm}, GLM-4~\cite{glm2024chatglm}, GPT-3.5-turbo (gpt-3.5-turbo-1106)~\cite{achiam2023gpt}, GPT-4o-mini (gpt-4o-mini-2024-07-18)~\cite{achiam2023gpt}, Qwen-2.5-72B-Instruct~\cite{bai2023qwen}, LLaMA-3.3-70B-Instruct~\cite{touvron2023llama}, and Claude-3.5-sonnet (claude-3-5-sonnet-20241022).
To reduce the risk of sampling variability, we set the temperature for all LLMs to 0.

We evaluated the performance of LLMs under three settings, simulating ideal and noisy scenarios:
\begin{itemize}[leftmargin=*]
\item \textbf{Zero Shot.}  The LLM generates answers without referencing legal knowledge, relying solely on its internal knowledge and reasoning abilities.

\item \textbf{Retriever.} The model generates answers using the top 5 documents retrieved by the retriever. In our experiments, we use the GTE-Qwen2-1.5B-instruct combined with query rewriting strategy, as this combination achieved the best recall rate.

\item \textbf{Reference.} The model generates answers with relevant legal articles annotated by legal experts. This evaluates the LLM's ability to solve the current issue under ideal knowledge conditions.

\end{itemize}

We use keyword accuracy and LLM judge scores as evaluation metrics.
Since legal terms often have unique meanings, a higher keyword accuracy indicates that the response covers more key legal knowledge.
In LLM-as-a-judge, we use the open-source LLM \textbf{Qwen-2.5-72B-Instruct} as the evaluator to ensure reproducibility.

\subsection{Generation Result}
Table \ref{genera} reports the performance of LexRAG under different LLMs and settings. Based on the experimental results, we have the following observations:

\begin{itemize}[leftmargin=*]
\item In terms of keyword accuracy, the performance under the reference setting is the best, followed by the retriever setting, while the zero-shot setting performs the worst. This indicates that current LLMs lack sufficient legal knowledge to generate relevant response.
When provided with relevant legal knowledge,  LLMs can generate responses that include more keywords.

\item Surprisingly, we observe that in the LLM judge score, the retriever setting does not consistently lead to performance improvements. In contrast, the reference setting consistently results in higher LLM judge scores. 
We believe this discrepancy occurs because when LLMs are provided with noisy or incomplete legal provisions, their limited legal knowledge prevents them from accurately referencing and analyzing the information, ultimately leading to lower scores.
These results suggest that advanced legal consultation systems cannot solely rely on retrieval techniques. To achieve optimal performance, it is crucial to also enhance the foundational LLM's understanding of legal concepts and reasoning.

\item Overall, we observe that Qwen-2.5-72B-Instruct achieved the best performance, followed by GLM-4. This may be due to the fact that these LLMs were developed by the Chinese community, which may make them better suited to legal consultation conversations in the Chinese legal domain.
However, even the best-performing LLMs still struggle to achieve a score of 8 in legal consultation scenarios. 
Given the complexity and precision required in the legal field, we recommend that the community focus on developing AI technologies specifically tailored to the unique needs and nuances of legal contexts.

\end{itemize}

\section{Limitation}

Although LexRAG advances the evaluation of RAG systems in the legal domain, there are still some limitations that need to be further addressed. First, LexRAG primarily focuses on Chinese legal scenarios, which limits its applicability in broader multilingual contexts. We plan to release an updated version supporting English in future iterations to expand its scope and enhance its cross-language evaluation capabilities. Second, due to the privacy and security constraints of real-world multi-turn consultation dialogues, the subsequent dialogue data in LexRAG is primarily annotated by legal experts. While this strategy ensures data quality and legality, it does not fully reflect the diversity and non-standardized interaction scenarios that may occur in real-world legal dialogues. To address this issue, future research will explore ways to leverage simulated data and artificial intelligence technologies while ensuring privacy protection, to better capture the complexity and demands of real-world multi-turn legal consultation conversations.

\section{Conclusion}
In this paper, we introduce LexRAG, a benchmark specifically designed to evaluate RAG systems in multi-turn legal consultation conversations. LexRAG comprises 1,013 consultations and 17,228 candidate legal articles, offering a comprehensive platform for assessing both conversational knowledge retrieval and response generation within the legal domain. In addition, we present LexiT, an open-source evaluation toolkit that provides a set of tools for automated, reproducible assessments of RAG systems in legal contexts. This toolkit enables detailed, fine-grained evaluations of various LLMs and retrieval methods, contributing to the advancement of AI applications in the legal field. In the future, we plan to develop RAG technologies more tailored to legal scenarios and expand LexRAG to support additional languages and legal systems, fostering the global advancement of intelligent judicial technologies.

\balance
\bibliographystyle{ACM-Reference-Format}
\bibliography{sample-base.bib}
\end{document}